\documentclass[10pt,twocolumn,letterpaper]{article}

\usepackage{cvpr}
\usepackage{times}
\usepackage{epsfig}
\usepackage{graphicx}
\usepackage{amsmath}
\usepackage{amssymb}
\usepackage{algpseudocode,algorithm,algorithmicx}
\usepackage{comment}

\usepackage[font=small,labelfont=bf]{caption} 
\usepackage{booktabs,siunitx}
\usepackage{tabularx}
\usepackage[usenames, dvipsnames]{xcolor}
\usepackage{multirow,bigdelim}
\usepackage{subcaption}
\usepackage{float}
\usepackage{enumitem}
\usepackage{textpos}

\graphicspath{{figures_pdf/}}

\newcommand{\B}[1]{{\textbf{#1}}}
\newcommand{\SC}[1]{{\textsc{#1}}}
\newcommand{\commentout}[1]{}

\newcommand{\refalg}[1]{Algorithm~\ref{#1}}
\newcommand{\refeqn}[1]{Equation~\ref{#1}}
\newcommand{\reffig}[1]{Fig.~\ref{#1}}
\newcommand{\reftbl}[1]{Table~\ref{#1}}
\newcommand{\refsec}[1]{Sec.~\ref{#1}}

\newcommand*\Let[2]{\State #1 $\gets$ #2}

\usepackage{amsmath}
\DeclareMathOperator*{\argmax}{arg\,max}

\usepackage{breakcites}
\definecolor{citecolor}{RGB}{34, 139, 34}
\usepackage[pagebackref=true,breaklinks=true,colorlinks,citecolor=citecolor,bookmarks=false,hypertexnames=false]{hyperref}

\cvprfinalcopy %
\def\cvprPaperID{XXXX} %

\ifcvprfinal\fi
\begin{document}

\title{\textsc{Ego-Topo}: Environment Affordances from Egocentric Video}

\author{
Tushar Nagarajan$^{1}$~~~
Yanghao Li$^{2}$~~~
Christoph Feichtenhofer$^{2}$~~~
Kristen Grauman$^{1,2}$ \\
$^{1}$ UT Austin~~~
$^{2}$ Facebook AI Research\\
\tt\small tushar@cs.utexas.edu, \{lyttonhao, feichtenhofer, grauman\}@fb.com
}

\maketitle

\begin{abstract}
First-person video naturally brings the use of a physical environment to the forefront, since it shows the camera wearer interacting fluidly in a space based on his intentions.  However, current methods largely separate the observed actions from the persistent space itself.  We introduce a model for environment affordances that is learned directly from egocentric video.  The main idea is to gain a human-centric model of a physical space (such as a kitchen) that captures (1) the primary spatial zones of interaction and (2) the likely activities they support.  Our approach decomposes a space into a topological map derived from first-person activity, organizing an ego-video into a series of visits to the different zones.  Further, we show how to link zones across multiple related environments (e.g., from videos of multiple kitchens) to obtain a consolidated representation of environment functionality.
On EPIC-Kitchens and EGTEA+, we demonstrate our approach for learning scene affordances and anticipating future actions in long-form video.
Project page: \url{http://vision.cs.utexas.edu/projects/ego-topo/}

\end{abstract}

\begin{textblock*}{\textwidth}(0cm,-16cm)
\centering
In Proceedings of the IEEE Conference on Computer Vision and Pattern Recognition (CVPR), 2020. 
\end{textblock*}

\begin{figure}[t!]
\centering
\includegraphics[width=\columnwidth]{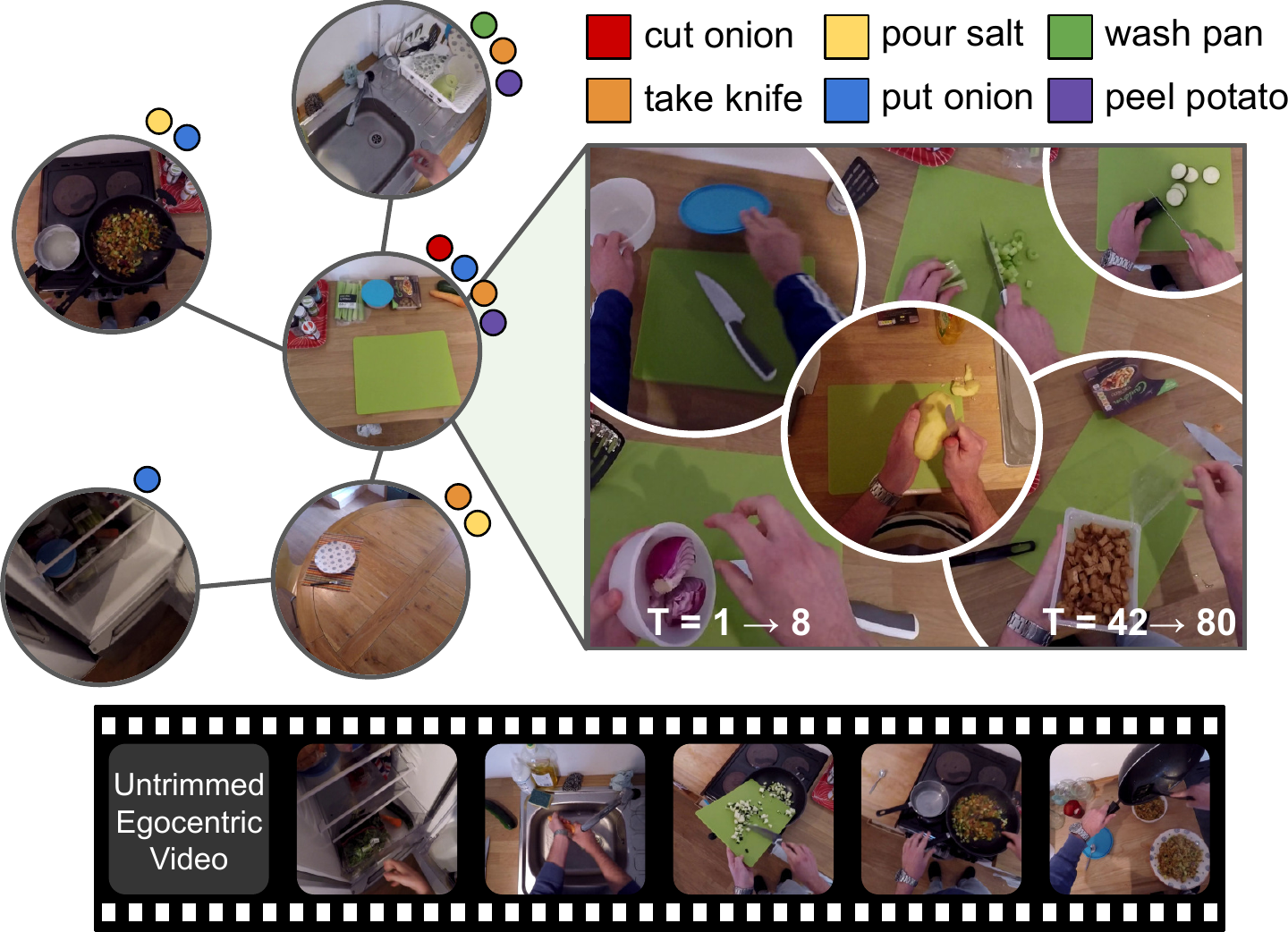}
\vspace*{-0.25in}
\caption{
\textbf{Main idea.}
Given an egocentric video, we build a topological map of the environment %
that reveals \emph{activity-centric} zones and the sequence in which %
they are visited. These maps capture 
the close tie between a physical space and how it is used by people, %
which we use to 
infer affordances of spaces (denoted here with color-coded dots) and anticipate future actions in long-form video.
}
\vspace*{-0.10in}
\label{fig:concept}
\end{figure}

\section{Introduction}

\emph{``The affordances of the environment are what it offers the animal, what it provides or furnishes... It implies the complementarity of the animal and the environment."}---James J. Gibson, 1979
\vspace{0.1in}

In traditional third-person images and video, we see a moment in time captured intentionally by a photographer who paused to actively record the scene.  As a result, scene understanding is largely about answering the \emph{who/where/what} questions of recognition: what objects are present? is it an indoor/outdoor scene? where is the person and what are they
doing?~\cite{quattoni2009recognizing,pandey2011scene,zhou2017places,lin2014microsoft,zhou2019semantic,johnson2015image,xu2017scene,gkioxari2018detecting}.

In contrast, in video captured from a first-person ``egocentric" point of view, we see the environment through the eyes of a person passively wearing a camera.  The %
surroundings are 
tightly linked to the camera-wearer's ongoing interactions with the environment.  As a result, scene understanding in egocentric video also entails \emph{how} questions: how can one use this space, now and in the future?
what areas are most conducive to a given activity? %

Despite this link between activities and environments, existing first-person video understanding models typically ignore that the underlying environment is a persistent physical space.  They instead treat the video as fixed-sized chunks of frames to be fed to neural networks~\cite{ma2016going,damen2018scaling,furnari2019rulstm,sudhakaran2019lsta,miech2019leveraging,li2018eye}.
Meanwhile, methods that do model the environment via dense geometric reconstructions~\cite{soo2016egocentric,guan2019generative,rhinehart2017first} suffer from SLAM failures---common in quickly moving head-mounted video---and do not discriminate between those 3D structures that are relevant to human actions and those that are not (\eg, a cutting board on the counter versus a random patch of floor).
We contend that neither the ``pure video" nor the ``pure 3D" perspective %
adequately captures the scene as an action-affording space.

Our goal is to build a model for an environment that captures how people use it.
We introduce an approach called \textsc{Ego-Topo} that converts %
egocentric video into a topological map consisting of activity ``zones" and their rough spatial proximity.  
Taking cues from Gibson's vision above, each zone is a region of the environment that \emph{affords a coherent set of interactions}, as opposed to a uniformly shaped region in 3D space.   See \reffig{fig:concept}.

Specifically, from egocentric video of people actively using a space, %
we link frames across time based on (1) the physical spaces they share and (2) the functions afforded by the zone, regardless of the actual physical location.  For example, for the former criterion, a dishwasher loaded at the start of the video is linked to the same dishwasher when unloaded, and to the dishwasher on another day.  For the latter,
a trash can in one kitchen could link to the garbage disposal in another: though visually distinct, both locations allow for the same action---discarding food. See \reffig{fig:linking}. 

In this way, we re-organize egocentric video into ``visits" to known zones, rather than a series of unconnected clips.  We show how %
doing so allows us to reason about first-person behavior (\eg, what are the most likely actions a person will do in the future?) and the environment itself (\eg, what are the possible object interactions that are likely in a particular zone, even if not observed there yet?).

Our \textsc{Ego-Topo} approach offers advantages over the existing models discussed above.  Unlike the ``pure video" approach, it provides a concise, spatially structured representation of the past.  Unlike the ``pure 3D" approach, our map is defined organically by people's use of the space.

We demonstrate our model on two key tasks: inferring likely object interactions in a %
novel view
and anticipating the actions needed to complete a long-term activity in first-person video.  These tasks illustrate how a vision system that can successfully reason about scenes' functionality would contribute to applications in augmented reality (AR) and robotics.  For example, an AR system that knows where actions are possible in the environment could interactively guide a person through a tutorial; a mobile robot able to learn from video how people use a zone would be primed to act without extensive exploration. 

On two challenging %
egocentric datasets, EPIC and EGTEA+, we show the value of modeling the environment explicitly for egocentric video understanding tasks, leading to more robust scene affordance models, and improving over state-of-the-art long range action anticipation models.

\section{Related Work}

\paragraph{Egocentric video}

Whereas the camera is a bystander in traditional third-person vision, in first-person or egocentric vision, the camera is worn by a person interacting with the surroundings firsthand.  %
This special viewpoint offers an array of interesting challenges, such as detecting gaze~\cite{li2013learning,huang2018predicting}, monitoring human-object interactions~\cite{cai2016understanding,damen2016you,nagarajan2018grounded}, creating daily life activity summaries~\cite{lu2013story,lee2015predicting,yonetani2016visual,lu2015personal}, or inferring the camera wearer's identity or body pose~\cite{hoshen2016egocentric,jiang2017seeing}.  The field is growing quickly in recent years, thanks in part to new ego-video benchmarks~\cite{damen2018scaling,li2018eye,pirsiavash2012detecting,sigurdsson2018charades}.

Recent work to recognize or anticipate actions in egocentric video adopts state-of-the-art video models from third-person video, like two-stream networks~\cite{li2018eye,ma2016going}, 3DConv models~\cite{damen2018scaling,pirri2019anticipation,miech2019leveraging}, or recurrent networks~\cite{furnari2019rulstm,gao2017red,shi2018action,sudhakaran2019lsta}.
In contrast, our model grounds first-person activity in a persistent topological encoding of the environment.  Methods that leverage SLAM together with egocentric video~\cite{guan2019generative,rhinehart2017first,soo2016egocentric} for activity forecasting also allow spatial grounding, though in a metric manner and with the challenges discussed above. 

\vspace*{-0.1in}
\paragraph{Structured video representations}

Recent work explores ways to enrich video representations with more structure.  Graph-based methods %
encode relationships between detected objects: nodes are objects or actors, and edges specify their spatio-temporal 
layout or semantic relationships (\eg, is-holding)~\cite{wang2018videos,baradel2018object,ma2018attend,zhang2019structured}.
Architectures for composite activity 
aggregate action primitives across
the video~\cite{girdhar2017actionvlad,hussein2019timeception,hussein2019videograph}, memory-based models record a recurrent network's state%
~\cite{pirri2019anticipation}, and 3D convnets augmented with long-term feature banks provide temporal context~\cite{wu2019long}.
Unlike any of the above, our approach encodes video in a human-centric manner according to how people use a space.  In our graphs, nodes are spatial zones and connectivity depends on a person's visitation over time.

\vspace*{-0.1in}
\paragraph{Mapping and people's locations}

Traditional maps use simultaneous localization and mapping (SLAM) to obtain dense metric measurements, viewing a space in strictly geometric terms.  Instead, recent %
work in embodied visual navigation explores learned maps that leverage both visual patterns as well as geometry, with the advantage of extrapolating to novel environments (\eg, \cite{gupta2017unifying,gupta2017cognitive,savinov2018semi,henriques2018mapnet,fang2019scene}).  Our approach shares this motivation. %
However, unlike any of the above, our approach analyzes egocentric video, as opposed to controlling a robotic agent.  Furthermore, whereas existing maps are derived from a robot's exploration, our maps are derived from \emph{human behavior}.  %

Work in ubiquitous computing tracks people %
to see where they spend time in an environment~\cite{koile2003activity,ashbrook2002learning}, and ``personal locations"  manually specified by the camera wearer (\eg, my office) can be recognized using supervised learning~\cite{furnari2018personal}.  In contrast, our approach automatically discovers zones of activity from ego-video, and it links action-related zones across multiple environments.

\vspace*{-0.1in}
\paragraph{Affordances}
Whereas we explore the affordances of \emph{environments}, prior work largely focuses on \emph{objects},
where the goal is to anticipate how an object can be used---e.g., learning to model object manipulation~\cite{alayrac2017joint,cai2016understanding}, how people would grasp an object~\cite{koppula2014physically,nagarajan2018grounded,fang2018demo2vec,damen2016you},  
or how body pose benefits object recognition~\cite{delaitre2012scene,grabner2011makes}.  The affordances of scenes are less studied.  Prior work explores how a third-person view of a scene suggests likely 3D body poses that would occur there~\cite{savva2014scenegrok,wang2017binge,gupta20113d} and vice versa~\cite{fouhey2014people}.  More closely related to our work, Action Maps~\cite{rhinehart2016learning} estimate missing activity labels for regular grid cells in an environment, using matrix completion with object and scene similarities as side information. In contrast, our work considers affordances not strongly tied to a single object's appearance, %
and we introduce a graph-based video encoding derived from our topological maps that benefits action %
anticipation.

\section{\textsc{Ego-Topo} Approach}

We aim to organize egocentric video %
into a map of activity ``zones"---%
regions that afford a coherent set of interactions---and ground the video as a series of visits to these zones. 
Our \textsc{Ego-Topo} 
representation offers
a middle ground between the ``pure video'' and ``pure 3D'' approaches %
discussed above, which
either ignore the underlying environment by treating video as fixed-sized chunks of frames, or sacrifice important semantics of human behavior
by densely reconstructing the whole environment. 
Instead, our model %
reasons jointly about the environment and the agent: which parts of the environment are most relevant for human action, %
what interactions does each zone afford.  

Our approach is best suited to long term activities in egocentric video where zones are repeatedly visited and used in multiple ways over time. This definition applies broadly to common household and workplace environments (\eg, office, kitchen, retail store, grocery).
In this work,
we study kitchen environments using two public ego-video datasets (EPIC~\cite{damen2018scaling} and EGTEA+~\cite{li2018eye}), since cooking activities entail frequent human-object interactions and repeated use of multiple zones. %
Our approach is not intended for third-person video, short video clips, or video where the environment is constantly changing (\eg, driving down a street).

Our approach first trains a zone localization network to discover commonly visited spaces from egocentric video %
(\refsec{sec:Approach1}). Then, given a novel video, we use the network to assign video clips to zones %
and create a topological map (graph) for the environment.  We further link zones based on their function across video instances to create consolidated maps (\refsec{sec:Approach2}). Finally, we 
leverage the resulting graphs
to uncover environment affordances %
(\refsec{sec:Approach3}) and anticipate future actions in long videos (\refsec{sec:Approach4}).

\subsection{Discovering Activity-Centric Zones} \label{sec:Approach1}

We %
leverage egocentric video of human activity %
to discover important ``zones'' for action. 
At a glance, one might attempt to discover spatial zones based on visual clustering or geometric partitions.  However, 
 clustering visual features (\eg, from a pretrained CNN) is insufficient since 
manipulated objects often feature prominently in ego-video, making the features sensitive to the set of objects present. For example, a sink with a cutting-board being washed vs.~the same sink at a different time filled with plates would cluster into different zones. %
On the other hand, SLAM localization is often unreliable
due to quick motions characteristic of egocentric video.\footnote{For example, on the EPIC Kitchens dataset, only $44\%$ of frames %
can be accurately registered with a state-of-the-art SLAM algorithm~\cite{mur2015orb}.}
Further, SLAM reconstructs all parts of the environment indiscriminately, without regard for their ties to human action or lack thereof, \eg, giving the same capacity to a kitchen sink area as it gives to a random wall.

\begin{figure}[t]
\centering
\includegraphics[width=\columnwidth]{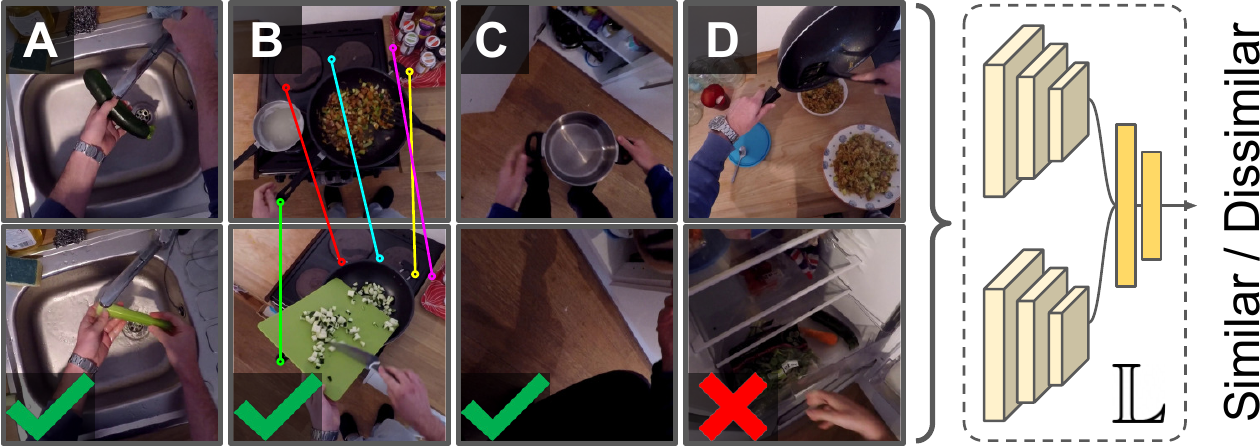}
\vspace*{-0.25in}  %
\caption{\textbf{Localization network.} Our similarity criterion goes beyond simple visual similarity (A), allowing our network to recognize the stove-top area (despite dissimilar features of prominent objects) with a consistent homography (B), or the seemingly unrelated views at the cupboard that are temporally adjacent (C), while distinguishing between dissimilar views sampled far in time (D).}
\vspace*{-0.10in}
\label{fig:locnetwork}
\end{figure}

To address these issues, we propose a zone discovery procedure that links views based on both their visual content and their visitation by the camera wearer.
The basis for this procedure is a
localization network that estimates the similarity of a pair of video frames, designed as follows.

We sample pairs of frames from videos that are segmented into a series of action clips. Two training frames are %
similar if (1) they are near in time (separated by fewer than 15 frames) or from the same action clip, or (2) there are at least 10 inlier keypoints consistent with their estimated homography.
The former allows us to capture the spatial coherence revealed by the person's 
tendency to dwell by action-informative zones, while the latter allows us to capture repeated backgrounds despite significant foreground object changes.
Dissimilar frames are temporally distant views with low visual feature similarity, or incidental views in which no actions occur.
See Fig.~\ref{fig:locnetwork}.
We use SuperPoint~\cite{detone2018superpoint} keypoint descriptors to estimate homographies, and euclidean distance between pretrained ResNet-152~\cite{he2016deep} features for visual similarity. 

The sampled pairs are used to train $\mathbb{L}$, a Siamese network with a ResNet-18~\cite{he2016deep} backbone, followed by a 5 layer multi-layer perceptron (MLP)%
, using cross entropy to predict whether the pair of views is similar or dissimilar. 
The network predicts the probability $\mathbb{L}(f_t, f_t')$ that two frames $f_t, f_t'$ in an egocentric video belong to the same zone.   

Our localization network draws inspiration from the retrieval network employed in~\cite{savinov2018semi} to build maps for embodied agent navigation, and more generally prior work leveraging temporal coherence to self-supervise image similarity~\cite{hadsell2006dimensionality,mobahi2009deep,jayaraman2016slow}. 
However, whereas the network in~\cite{savinov2018semi} is learned from view sequences generated by a randomly navigating agent, ours learns from ego-video taken by a human acting purposefully in an environment rich with object manipulation. In short, nearness in~\cite{savinov2018semi} is strictly about physical reachability, whereas nearness in our model is about human interaction in the environment.

\begin{algorithm}[t]
\small
  \begin{algorithmic}[1] 
    \caption{Topological affordance graph creation.}
    \label{alg:topomap}
    \Require{A sequence of frames $(f_1, ... f_T)$ of a video}
    \Require{Trained localization network $\mathbb{L}$ (Sec.~\ref{sec:Approach1})}
    \Require{Node similarity threshold $\sigma$ and margin $m$}
    \State Create a graph $G = (N, E)$ with node $n_1 = \{(f_1\rightarrow f_1)\}$
      \For{$t \gets 2 \textrm{ to } T $}
        \Let {$s^*$}{$\max_{n \in N} s_f(f_t, n)$ --- \refeqn{eq:sim}}  %
        \If{$s^* > \sigma$}
            \State Merge $f_t$ with node  $n^*=\argmax_{n \in N} s_f(f_t,n)$
            \State If $f_t$ is a consecutive frame in $n^*$: Extend last visit $v$
            \State Else: Make new visit $v$ with $f_t$
        \ElsIf {$s^* < \sigma - m$}
            \State Create new node, add visit with $f_t$, and add to $G$
        \EndIf
        \State Add edge from last node to current node
      \EndFor 
    \Ensure{\textsc{Ego-Topo} topological affordance graph $G$ per video}
  \end{algorithmic}
\end{algorithm}

\subsection{Creating the Topological Affordance Graph}\label{sec:Approach2}

With a trained localization network, we %
process the stream of frames in a new untrimmed, unlabeled egocentric video to build a topological map of its environment.  For a video $\mathcal{V}$ with $T$ frames $(f_1, ..., f_T)$, we create a graph $G = (N, E)$ with nodes $N$ and edges $E$. Each node of the graph is a zone and %
records a collection of ``visits''---%
clips
from the egocentric video at that location. For example,  a cutting board counter visited at $t=1$ and $t=42$, for 7 and 38 frames each, will be represented by a node $n \in N$ with visits 
$\{v_1 = (f_1 \rightarrow f_{8}), v_2 = (f_{42} \rightarrow f_{80})\}$.
See \reffig{fig:concept}. 

We initialize the graph with a single node $n_1$ corresponding to a visit with just the first frame. For each subsequent frame $f_t$, we compute the average \emph{frame-level} similarity score $s_f$ for the frame compared to each of the nodes $n \in N$ using the localization network from \refsec{sec:Approach1}: 
\begin{align}
  s_f(f_t, n) &= \frac{1}{|n|}\sum_{v \in n} \mathbb{L}(f_t, f_v) \\
  s^* &= \max_{n \in N} ~ s_f(f_t, n),  \label{eq:sim}
\end{align}       
where $f_v$ 
is the center frame selected from each visit $v$ in node $n$. If the network is confident that the frame is similar to one of the nodes, it is merged with the highest scoring node $n^*$ corresponding to $s^*$. Alternately, if the network is confident that this is a new location (very low $s^*$), a new node is created for that location, and an edge is created from the previously visited node. The frame is ignored if the network is uncertain about the frame.
\refalg{alg:topomap} summarizes the construction algorithm. 
See Supp.~for more details.

%
%

%

%
%
%

%
%
%
%
%
%
%
%
%
%
%
%
%
%
%
%
%
%
%
%
%
%
%

%
%
%
%
%
    
%
%
\begin{figure}[t]
\centering
\includegraphics[width=\columnwidth]{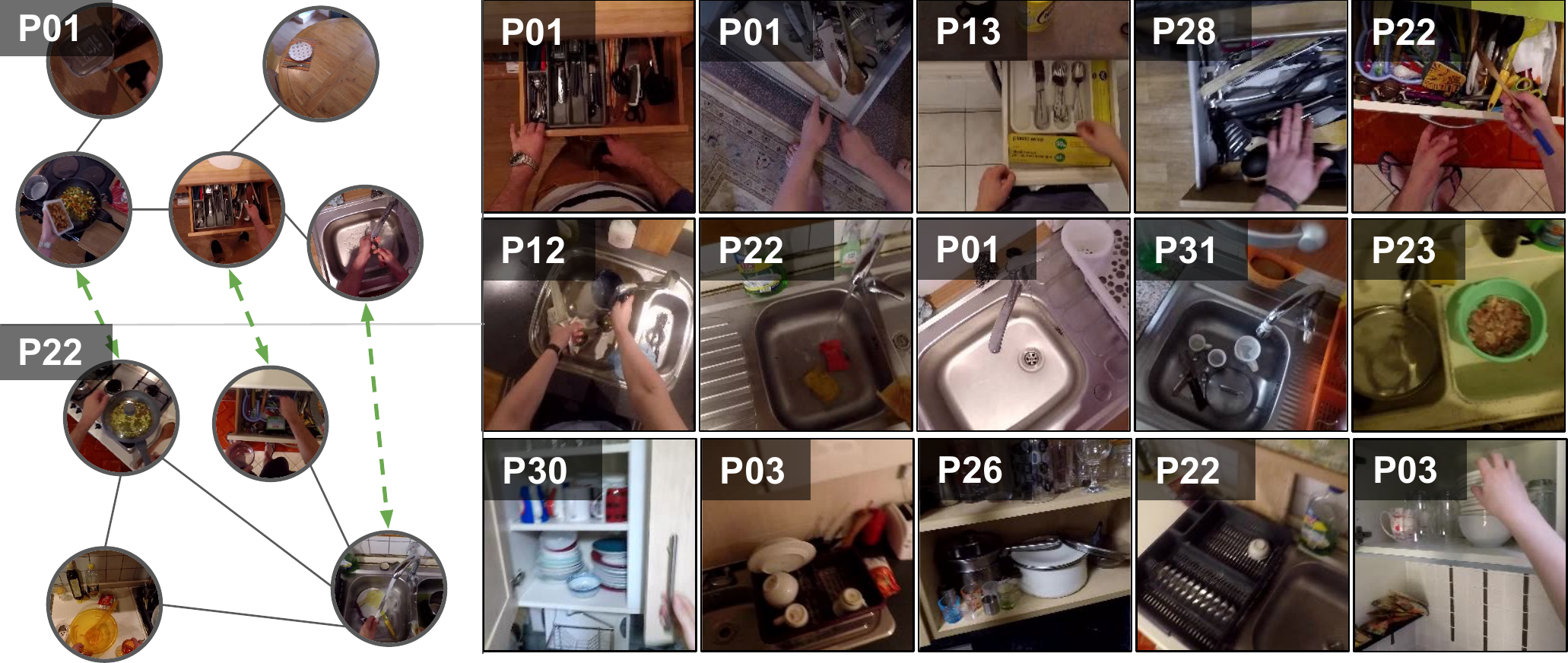}
\vspace*{-0.25in}
\caption{\textbf{Cross-map linking.} Our linking strategy aligns multiple kitchens (P01, P13 etc.) by their common spaces (\eg, drawers, sinks in rows 1-2) and visually distinct, but functionally similar spaces (\eg, dish racks, crockery cabinets in row 3). %
}
\vspace*{-0.10in}
\label{fig:linking}
\end{figure}

When all frames are processed, we are left with a graph of the environment per video where nodes correspond to zones where actions take place (and a list of visits to them) and the edges capture weak spatial connectivity between zones based on how people traverse them.  %

Importantly, beyond per-video maps, our approach also creates cross-video and cross-environment maps that link spaces by their function.
We show how to link zones across 1) multiple episodes in the same environment and 2) multiple environments with shared functionality.
To do this, for each node $n_i$ we use a pretrained action/object classifier to compute $(\mathbf{a}_i, \mathbf{o}_i)$, the distribution of actions and active objects\footnote{An active object is an object involved in an interaction.
}
that occur in all visits to that node. We then compute a \emph{node-level} functional similarity score:
\begin{equation}
s_n(n_i, n_j) = -\frac{1}{2} \left ( KL(\mathbf{a}_i||\mathbf{a}_j) + KL(\mathbf{o}_i||\mathbf{o}_j) \right ), \label{eq:kld}
\end{equation}
where KL is the KL-Divergence. 
We score pairs of nodes across all kitchens, and perform hierarchical agglomerative clustering to link nodes with functional similarity.  %
%
%
%
%

\begin{figure*}[htb!]
\centering
\includegraphics[width=\linewidth]{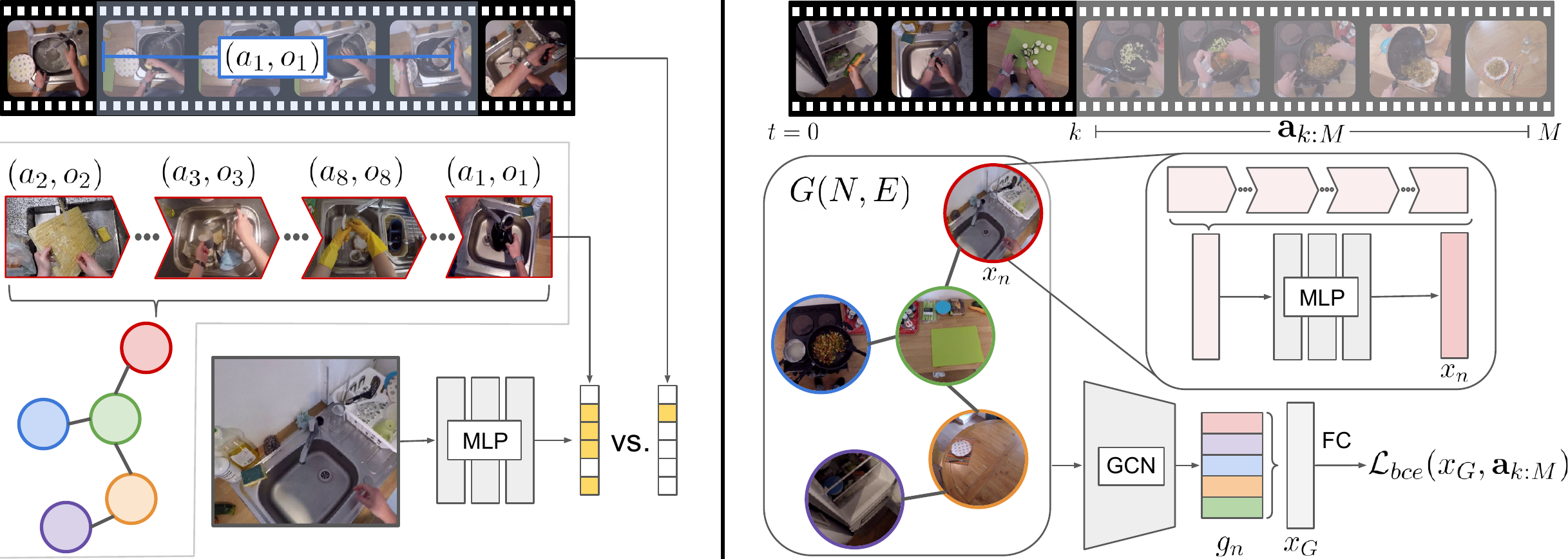}
\caption{\textbf{Our methods for environment affordance learning (L) and long horizon action anticipation (R)}. 
\textbf{Left panel}: Our \textsc{Ego-Topo} %
graph allows multiple affordance labels to be associated with visits to zones, compared to single action labels in annotated video clips. Note that these visits %
may come from different videos of the same/different kitchen---which provides a more robust view of affordances (cf.~\refsec{sec:Approach3}).
\textbf{Right panel}: We use our topological graph to aggregate features for each zone and consolidate information across zones via graph convolutional operations, to create a concise video representation for long term video anticipation (cf.~\refsec{sec:Approach4}).
}
\label{fig:tasks}
\vspace{-0.1in}
\end{figure*}

Linking nodes in this way offers several benefits. First, not all parts of the kitchen are visited in every episode (video). We link zones across different episodes in the \emph{same} kitchen to create a \emph{combined} map of that kitchen that accounts for the persistent physical space underlying multiple video encounters.
Second, we link zones \emph{across} kitchens to create a \emph{consolidated} kitchen map, which reveals how different kitchens relate to each other. 
For example, a gas stove in one kitchen could link to a hotplate in another, despite being visually dissimilar (see \reffig{fig:linking}).   Being able to draw such parallels is valuable when %
planning to act in a new unseen environment, as we will demonstrate below.

\subsection{Inferring Environment Affordances}\label{sec:Approach3}

Next, we %
leverage the proposed topological graph %
to predict a zone's affordances---\emph{all likely} interactions possible at that zone. 
Learning scene affordances is especially important when an agent must use a previously unseen environment to perform a task.  Humans seamlessly do this, \eg, cooking a meal in a friend's house; we are interested in AR systems and robots that learn to do so by watching humans.
We know that egocentric video of people performing daily activities %
reveals how different parts of the %
space are used.  %
Indeed, the actions observed per zone partially reveal its affordances.
However, since each clip of an ego-video shows a zone being used only for a single interaction, %
it falls short of capturing \emph{all likely} interactions at that location.

To overcome this limitation, our key insight is that %
linking zones within/across environments allows us to extrapolate labels for \emph{unseen interactions} at \emph{seen zones}, resulting in a more complete picture of affordances. In other words, having seen an interaction $(a_i, o_i)$ at a zone $n_j$ allows us to augment training for the affordance of $(a_i, o_i)$ at zone $n_k$, if zones $n_j$ and $n_k$ are functionally linked. See \reffig{fig:tasks} (Left).

To this end, we treat the affordance learning problem as a multi-label classification task that maps image features $x_i$ to an $A$-dimensional binary indicator vector $\mathbf{y}_i \in \{0,1\}^A$, where $A$ is the number of possible interactions. We generate training data for this task using the topological affordance graphs $G(N, E)$ defined in~\refsec{sec:Approach2}.

Specifically, we calculate \emph{node-level} affordance labels $\mathbf{y}_n$ for each node $n \in N$:
\begin{equation}
\mathbf{y}_n(k)= 1 ~~~~~ \text{for} ~~~ k \in \bigcup_{v \in n} \mathcal{A}(v),
\end{equation}
where %
$\mathcal{A}(v)$ is the set of all interactions that occur during visit $v$.\footnote{For consolidated graphs, $N$ refers to nodes \emph{after} clustering by Eq. 3.} Then, for each visit to a node $n$, we sample a frame, generate its features $x$, and use $\mathbf{y}_n$ as the  multi-label affordance target. %
We use a 2-layer MLP for the affordance classifier, followed by a linear classifier and sigmoid.  %
The network is trained using binary cross entropy loss. 
%

At test time, given an image $x$ %
in an environment,
this classifier directly predicts its
affordance probabilities.  See \reffig{fig:tasks} (Left). 
Critically, linking frames into zones and linking zones between environments %
allows us to share labels across instances in a manner
that benefits affordance learning, better than models that link data purely based on geometric or visual nearness (cf.~Sec.~\ref{sec:affordance_exps}).
 %
%
%

%

%

%
%

%
%
%
%
%

%
%
%

%

%

\subsection{Anticipating Future Actions in Long Video}\label{sec:Approach4}

Next, we leverage our topological affordance graphs for long horizon anticipation.  
In the anticipation task, we see a fraction of a long video (\eg, the first 25\%), and from that we must predict what actions will be done in the future. Compared to affordance learning, which benefits from how zones are functionally related to enhance static image understanding, %
long range action anticipation is a video understanding task that 
leverages how zones are laid out, and where actions are performed, to anticipate human behavior.

Recent action anticipation work~\cite{furnari2017next,zhou2015temporal,furnari2019rulstm,damen2018scaling,pirri2019anticipation,gao2017red,shi2018action} predicts the immediate next action (\eg in the next 1 second) rather than all future actions, for which an encoding of recent video information is sufficient. For long range anticipation, models need to understand how much progress has been made on the composite activity so far, and anticipate what actions need to be done in the future to complete it. For this, a structured representation of all past activity and affordances is essential. Existing long range video understanding methods~\cite{hussein2019timeception,hussein2019videograph,wu2019long} build complex models 
to aggregate information from the past, but do not model the environment explicitly, which we hypothesize is important for anticipating actions in long video. Our graphs provide a concise representation of observed activity, grounding frames in the spatial environment.  

Given an untrimmed video $\mathcal{V}$ with $M$ interaction clips %
each involving an action $\{a_1,...,a_M\}$ with some object, we see the first $k$ clips
and predict the future action labels as a $D$-dimensional binary vector $\mathbf{a_{k:M}}$, where $D$ is the number of action classes and $a_{k:M}^d = 1$ for $d \in \{a_{k+1},...,a_M\}$.

We generate the corresponding topological graph $G(N, E)$ built up to $k$ clips, and extract features $x_n$ for each node using a 2-layer MLP, over the average of clip features sampled from visits to that node.

Actions at one node influence future activities in other nodes. To account for this, we enhance node features by integrating neighbor node information from %
the topological graph using a graph convolutional neural network (GCN)~\cite{kipf2016semi}
\begin{equation}
g_n = ReLU \left ( \sum_{n' \in \mathcal{N}_n} W^Tx_{n'} + b  \right ),
\end{equation}
where $\mathcal{N}_n$ are the neighbors of node $n$, and $W, b$ are learnable parameters of the GCN.

The updated GCN representation $g_n$ for each individual node is enriched with global scene context from neighboring nodes, allowing patterns in actions across locations to be learned. For example, vegetables that are \emph{taken out} of the fridge in the past are likely to be \emph{washed} in the sink later. The GCN node features are then averaged to derive a representation of the video $x_G = \frac{1}{|N|} \sum_{n \in N} g_{n}$.  This is then fed to a linear classifier followed by a sigmoid to predict future action probabilities, trained using binary cross entropy loss, $\mathcal{L}_{bce}(x_G,\  \mathbf{a_{k:M}})$.

 At test time, given an untrimmed, unlabeled video %
showing the onset of a long composite activity,
our model can predict the actions that will likely occur in the future to complete it. %
See~\reffig{fig:tasks} (Right) and Supp.  
%
As we will see in results, grounding ego-video in the real environment---rather than treat it as an arbitrary set of frames---provides a stronger video representation for anticipation.

\section{Experiments}  \label{sec:exp}

We evaluate the proposed topological %
graphs for scene affordance learning and action anticipation in long videos.

\vspace{0.05in}
\noindent\textbf{Datasets}. \label{sec:datasets} We use two egocentric video datasets: %

\begin{itemize}[leftmargin=*]
\item \textbf{EGTEA Gaze+}~\cite{li2018eye} contains videos of 32 subjects following 7 recipes in a single kitchen. Each video captures a complete dish being prepared (\eg, potato salad, pizza), with clips annotated for interactions (\eg, open drawer, cut tomato), spanning 53 objects and 19 actions.

\item \textbf{EPIC-Kitchens}~\cite{damen2018scaling} contains videos of daily kitchen activities, and is not limited to a single recipe. It is annotated for interactions spanning 352 objects and 125 actions. Compared to EGTEA+, EPIC is larger, unscripted, and collected across multiple kitchens.
\end{itemize}

The kitchen environment 
has been the subject of several recent egocentric datasets~\cite{damen2018scaling,li2018eye,kuehne2014language,stein2013combining,rohrbach2012database,zhou2018towards}. Repeated interaction with different parts of the kitchen during complex, multi-step cooking activities is a rich %
domain for learning affordance and anticipation models.

\subsection{\textsc{Ego-Topo} for Environment Affordances} \label{sec:affordance_exps}

In this section, we evaluate how linking actions in zones and across environments can benefit affordances. %

\noindent\textbf{Baselines}. We compare the following methods:
\begin{itemize}[leftmargin=*]
\itemsep0em 

\item \textbf{\SC{ClipAction}} 
%
is a frame-level action recognition model trained to predict a single interaction label, given a frame from a video clip showing that interaction.

\item \textbf{\SC{ActionMaps}}~\cite{rhinehart2016learning} estimates affordances of locations via matrix completion with side-information. It assumes that nearby locations with similar appearance/objects have similar affordances. See Supp. for details.

\item \textbf{\SC{SLAM}} trains an affordance classifier with the same architecture as ours, and treats all frames from 
the same grid cell on the ground plane as positives for actions observed at any time in that grid cell.  $(x, y)$ locations are obtained from monocular SLAM~\cite{mur2015orb}, and cell size is based on the typical scale of an interaction area~\cite{guan2019generative}.  %
It shares our insight to link actions in the same location, but is limited to a uniformly defined location grid
and cannot link different environments. See Supp for details.

\item \textbf{\SC{KMeans}} clusters action clips using their visual features alone. We select as many clusters as there are nodes in our consolidated graph to ensure fair comparison.

\item \textbf{\SC{Ours}} 
We show the three variants from~\refsec{sec:Approach2} which use maps built from a single video (\SC{Ours-S}), multiple videos of the same kitchen (\SC{Ours-M}), and a functionally linked, consolidated map across kitchens (\SC{Ours-C}).

\end{itemize}

Note that all methods use the clip-level annotated data, in addition to data from linking actions/spaces.  They see the same video frames during training, only they are organized and presented with labels according to the method. 

\noindent\textbf{Evaluation}. We crowd-source annotations for afforded interactions. Annotators label a frame $x$ from the video clip with \emph{all} likely interactions at that location regardless of whether the frame shows it (\eg, turn-on stove, take/put pan etc.~at a stove), which is encoded as an $A$-dimensional binary target $\mathbf{y}$.

%
We collect 1020 instances spanning $A = 75$ interactions on EGTEA+ and 1155 instances over $A = 120$ on EPIC (see Supp.~for details).
All methods are evaluated on this test set.
We evaluate multi-label classification performance using mean average precision (mAP) over all afforded interactions, and separately for the rare and frequent ones ($<$10 and $>$100 training instances, respectively). 

\begin{table}[]
\resizebox{\columnwidth}{!}{
\begin{tabular}{|l|l|ll|l|l|ll|}
\multicolumn{1}{c}{}  &      \multicolumn{3}{c}{EPIC}        &  \multicolumn{1}{c}{} &    \multicolumn{3}{c}{EGTEA+} \\ 
\cline{1-4} \cline{6-8}

mAP $\rightarrow$     & \SC{All} & \SC{Freq}   & \SC{Rare}    &     & \SC{All}    & \SC{Freq}    & \SC{Rare}  \\ 
\cline{1-4} \cline{6-8} 

\SC{ClipAction}       & 26.8        & 49.7         & 16.1     &     & 46.3     & 58.4        & 33.1           \\
\SC{ActionMaps}~\cite{rhinehart2016learning}
                      & 21.0        & 40.8         & 13.4     &     & 43.6     & 52.9        & 31.3           \\
\SC{SLAM}             & 26.6        & 48.6         & 17.6     &     & 41.8     & 49.5        & 31.8           \\
\SC{KMeans}           & 26.7        & 50.1         & 17.4     &     & 49.3     & \B{61.2}    & 35.9           \\
\SC{Ours-S}           & 28.6        & 52.2         & 19.0     &     & 48.9     & 61.0        & 35.3           \\
\SC{Ours-M}           & 28.7        & 53.3         & 18.9     &     & \B{51.6} & \B{61.2}    & \B{37.8}       \\
\SC{Ours-C}           & \B{29.4}    & \B{54.5}     & \B{19.7} &     & --       & --          & --             \\ 
\cline{1-4} \cline{6-8} 
\end{tabular}
}
\caption{\textbf{Environment affordance prediction.}  Our method outperforms all other methods. 
Note that videos in EGTEA+ are from the same kitchen, and do not allow cross-kitchen linking. Values are averaged over 5 runs.}
\label{tbl:affordance}
\vspace{-0.1in}
\end{table}

\reftbl{tbl:affordance} summarizes the results. By capturing the persistent environment in our discovered zones and linking them across environments, our method outperforms all other methods on the affordance prediction task. All models perform better on EGTEA+, which %
has fewer interaction classes, contains only one kitchen, and has %
at least 30 training examples per afforded action (compared to EPIC where 10\% of the actions have a single annotated clip).

\begin{figure}[t]
\centering
\includegraphics[width=\columnwidth]{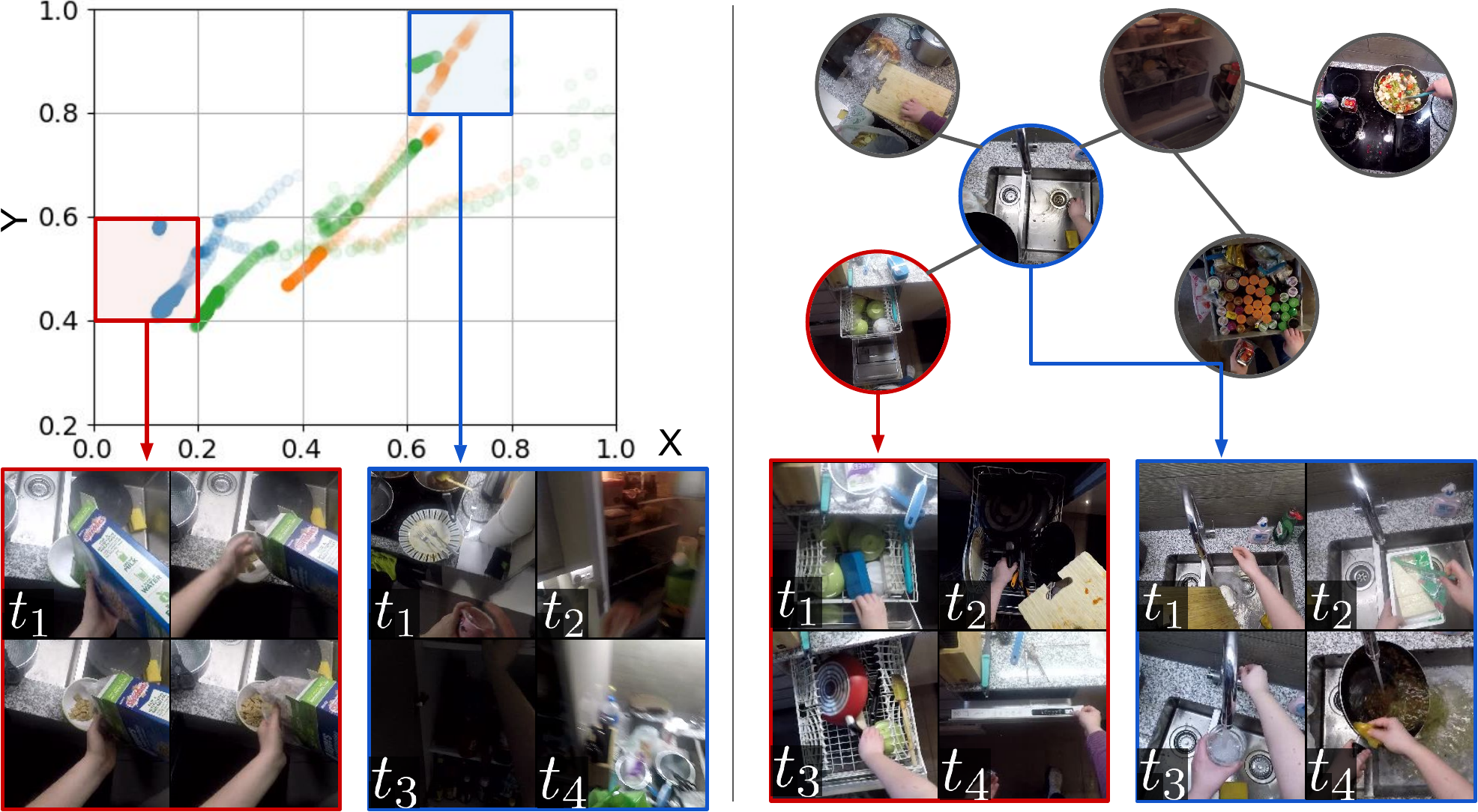}
\vspace*{-0.25in}
\caption{\textbf{SLAM grid vs graph nodes.} The boxes show frames from video that are linked to grid cells in the SLAM map (Left) and nodes in our topological map (Right). See text. %
}
\vspace*{-0.10in}
\label{fig:slam_vs_graph}
\end{figure}

\SC{SLAM} and \SC{ActionMaps}~\cite{rhinehart2016learning} rely on monocular SLAM, which introduces certain limitations. See \reffig{fig:slam_vs_graph} (Left). A single grid cell in the SLAM map reliably registers only small windows of smooth motion, often capturing only single action clips at each location. %
In addition, inherent scale ambiguities %
and uniformly shaped cells can result in incoherent activities placed in the same cell.
Note that this limitation stands even if SLAM were perfect.  Together, these factors hurt performance on both datasets, more severely affecting EGTEA+ due to the scarcity of SLAM data (only 6\% accurately registered). 
Noisy localizations also affect the kernel
computed by \SC{ActionMaps}, which accounts for physical nearness as well as similarities in object/scene features.
In contrast, a zone in our topological affordance graph corresponds to a coherent set of clips at different times, linking 
a more reliable and diverse set of actions, 
as seen in \reffig{fig:slam_vs_graph} (Right). 

Clustering using purely visual features in \SC{KMeans} helps consolidate information in EGTEA+ where all videos are in the same kitchen, but hurts performance where visual features are %
insufficient to capture coherent zones.

\textsc{Ego-Topo}'s linking of actions to discovered zones 
yields consistent improvements on both datasets. Moreover, aligning spaces based on function in the consolidated graph (\SC{Ours-C}) provides the largest improvement, especially for rare classes that may only be seen tied to a single location.  

\begin{figure}[t]
\centering
\includegraphics[width=\columnwidth]{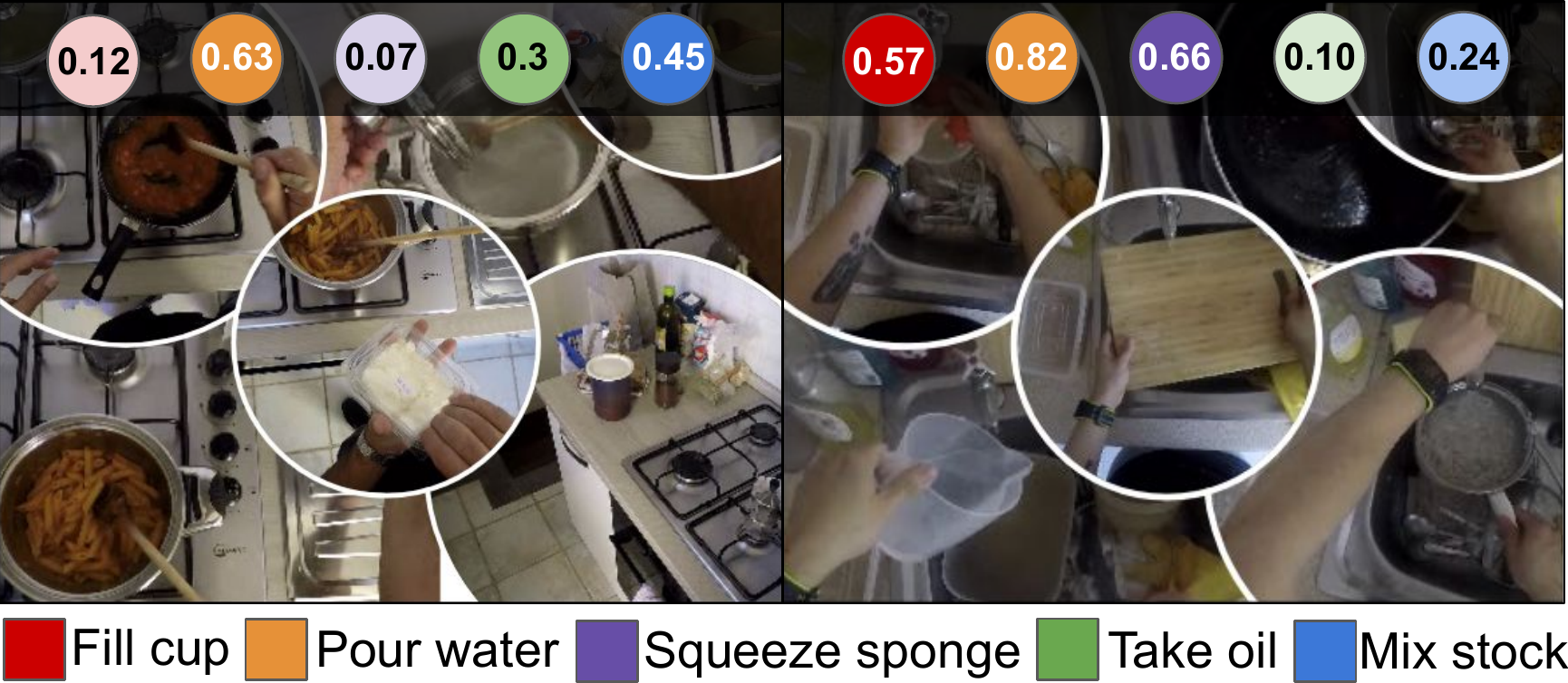}
\vspace*{-0.25in}
\caption{\textbf{Top predicted affordance scores for two graph nodes.} Our affordance model applied to node visits 
reveal zone affordances.  Images in circles are sampled frames from the two nodes.
}
\vspace*{-0.10in}
\label{fig:afford_qual}
\end{figure}

\reffig{fig:linking} and \reffig{fig:slam_vs_graph} show the diverse  actions captured in each node of our graph. Multiple actions at different times and from different kitchens are %
linked to the same %
zone, thus overcoming the sparsity in demonstrations and translating to a 
strong training signal for our scene affordance model. \reffig{fig:afford_qual}
shows example affordance predictions.%

\subsection{\textsc{Ego-Topo} for Long Term Action Anticipation} \label{sec:anticipation_exps}

Next we evaluate how the structure of our topological graph yields better video features for long term anticipation.

\noindent\textbf{Baselines}. We compare against the following methods:

\begin{itemize}[leftmargin=*]
\itemsep0em 

\item \textbf{\SC{TrainDist}} simply outputs the distribution of actions performed in all training videos, to test if a few dominant actions are repeatedly done, regardless of the video.

\item \textbf{\SC{I3D}} uniformly samples 64 clip features and averages them to generate a video feature. 

\item \textbf{\SC{RNN}} and \textbf{\SC{ActionVLAD}}~\cite{girdhar2017actionvlad} model temporal dynamics in video using LSTM~\cite{hochreiter1997long} layers and non-uniform pooling strategies, respectively.

\item \textbf{\SC{Timeception}}~\cite{hussein2019timeception} and \textbf{\SC{VideoGraph}}~\cite{hussein2019videograph} build complex temporal models using either multi-scale temporal convolutions or attention mechanisms over learned latent concepts from clip features over large time scales.

\end{itemize}

\begin{table}[]
\resizebox{\columnwidth}{!}{
\begin{tabular}{|l|l|ll|l|l|ll|}

\multicolumn{1}{c}{}    &      \multicolumn{3}{c}{EPIC} &  \multicolumn{1}{c}{} &  \multicolumn{3}{c}{EGTEA+} \\
\cline{1-4} \cline{6-8}

mAP $\rightarrow$       & \SC{All} & \SC{Freq} & \SC{Rare} &  & \SC{All}  & \SC{Freq} & \SC{Rare} \\ 
\cline{1-4} \cline{6-8} 
\SC{TrainDist}          & 16.5      & 39.1      & 5.7      &  & 59.1     & 68.2      & 35.2      \\
\SC{I3D}                & 32.7      & 53.3      & 23.0     &  & 72.1     & 79.3      & 53.3      \\
\SC{RNN}                & 32.6      & 52.3      & 23.3     &  & 70.4     & 76.6      & 54.3      \\
\SC{ActionVLAD}~\cite{girdhar2017actionvlad}
                        & 29.8      & 53.5      & 18.6     &  & 73.3     & 79.0      & 58.6  \\
\SC{VideoGraph}~\cite{hussein2019videograph}
                        & 22.5      & 49.4      & 14.0     &  & 67.7     & 77.1      & 47.2      \\
\SC{Timeception}~\cite{hussein2019timeception}
                        & 35.6	    & 55.9	    & 26.1    &  & \B{74.1}     & 79.7   & \B{59.7}      \\
\cline{1-4} \cline{6-8}
\SC{Ours w/o GCN}       & 34.6	    & 55.3	    & 24.9    &  & 72.5     & 79.5      & 54.2      \\ 
\SC{Ours}               & \B{38.0}	& \B{56.9}	& \B{29.2} &  & 73.5     & \B{80.7}  & 54.7      \\ 

\cline{1-4} \cline{6-8} 
\end{tabular}
}
\vspace{-0.05in}
\caption{\textbf{Long term anticipation results.} 
Our method outperforms all others on EPIC, and is best for many-shot classes on the simpler EGTEA+. Values are averaged over 5 runs. 
}
\label{tbl:anticipation}
\vspace{-0.15in}
\end{table}

%

The focus of our model is to generate a structured representation of past video. Thus, these methods that consolidate information over long temporal horizons are most appropriate for direct comparison. Accordingly, our experiments keep the anticipation module itself fixed (a linear classifier over a video representation), and vary the representation. Note that state-of-the-art anticipation models~\cite{furnari2019rulstm,abu2018will,ke2019time} ---which \emph{decode} future actions from such an encoding of past (observed) video---address an orthogonal problem, and can be used in parallel with our method.

\noindent\textbf{Evaluation}. $K$\% of each untrimmed video is given as input, and all actions in the future 100-$K$\% of the video must be predicted as a binary vector (does each action happen any time in the future, or not).
We sweep values of $K = [25\%, 50\%, 75\%]$ representing different anticipation horizons. %
We report multi-label classification performance as mAP over all action classes, and again in the low-shot (rare) and many-shot (freq) settings.

\begin{figure}[t]
\centering
\includegraphics[width=\columnwidth]{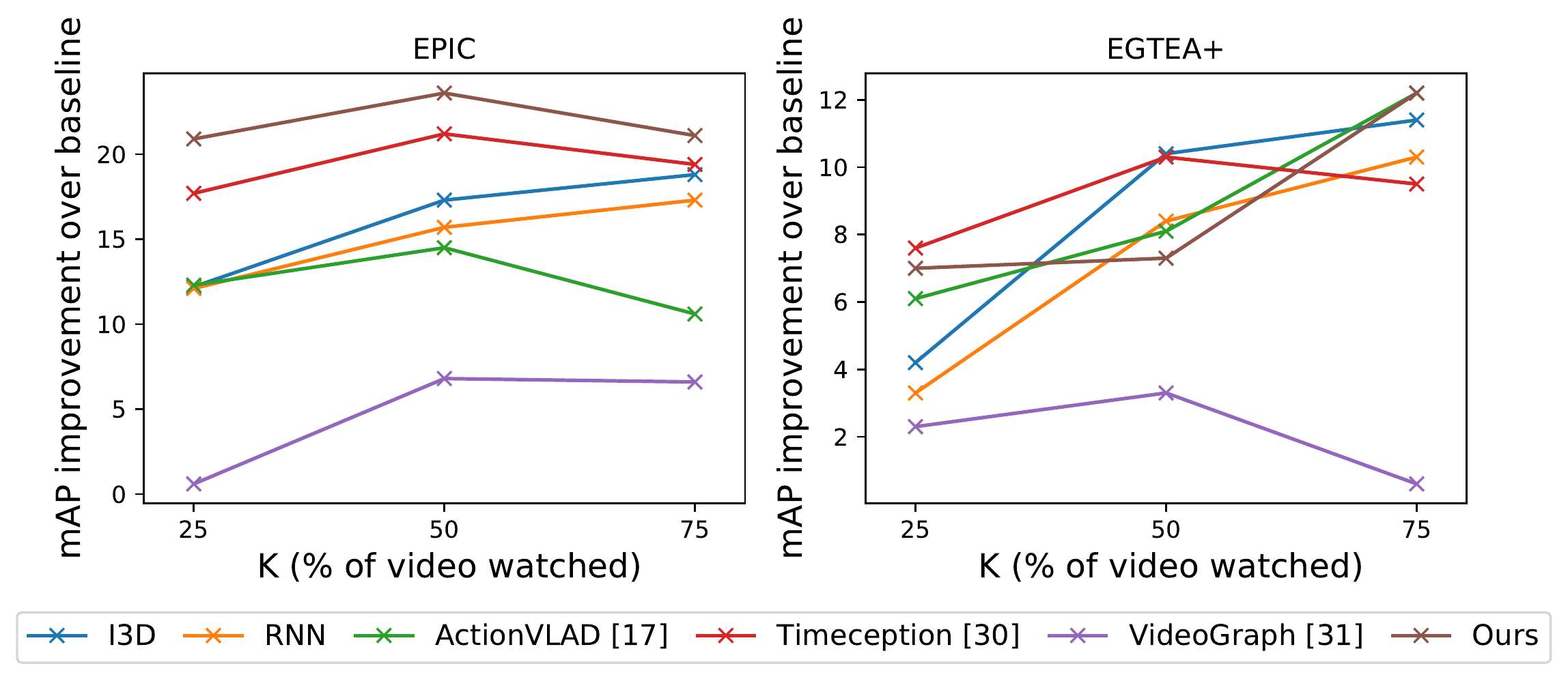}
\vspace*{-0.25in}
\caption{\textbf{Anticipation performance over varying prediction horizons.} $K\%$ of the video is observed, then the actions in the remaining $100-K\%$ must be anticipated. Our model outperforms all methods for all anticipation horizons on EPIC, and has higher relative improvements when predicting further into the future.
}%

\vspace*{-0.10in}
\label{fig:ant_by_K}
\end{figure}

\reftbl{tbl:anticipation} shows the results averaged over all $K$'s, and \reffig{fig:ant_by_K} plots results vs.~$K$. Our model outperforms all other methods on EPIC, improving over the next strongest baseline by 2.4\% mAP on all 125 action classes.  
On EGTEA+, our model matches the performance of models with complicated temporal aggregation schemes, and achieves the highest results for many-shot classes. 

EGTEA+ has a less diverse action vocabulary with a fixed set of recipes. \SC{TrainDist}, which simply outputs a fixed distribution of actions for every video, performs relatively well (59\% mAP) compared to its counterpart on EPIC (only 16.5\% mAP), highlighting that there is a core set of repeatedly performed actions in EGTEA+.

Among the methods that employ complex temporal aggregation schemes, \SC{Timeception} improves over \SC{I3D} on both datasets, though our method outperforms it on the larger EPIC dataset.
Simple aggregation of node level information (\SC{Ours w/o GCN}) still consistently outperforms most baselines.  However, including graph convolution is essential to outperform more complex models, which shows the benefit of encoding the physical layout and interactions between zones in our topological map.

\reffig{fig:ant_by_K} breaks down performance by anticipation horizon $K$.  %
On EPIC, our model is uniformly better across all prediction horizons, and it excels at predicting actions further into the future.
This highlights the benefit of our environment-aware video representation. On EGTEA+, our model outperforms all other models except \SC{ActionVLAD} on short range settings, but performs slightly worse at $K$=50\%.  On the other hand, \SC{ActionVLAD} falls short of all other methods on the more challenging EPIC data.

\begin{figure}[t]
\centering
\includegraphics[width=\columnwidth]{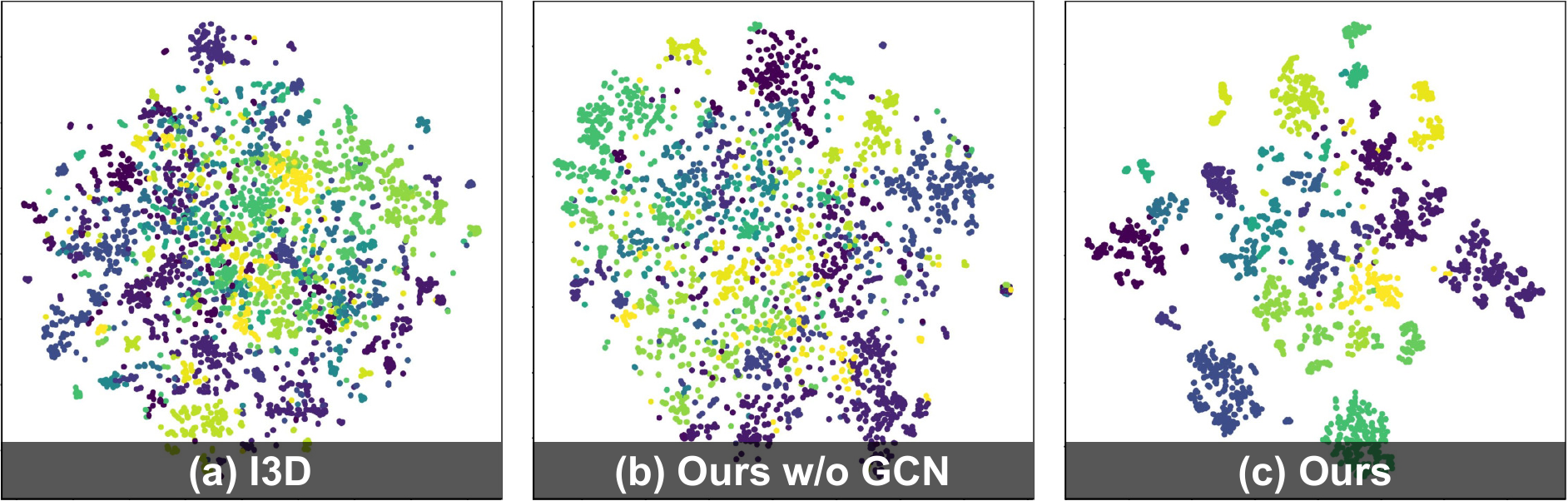}
\vspace*{-0.25in}
\caption{\textbf{t-SNE~\cite{maaten2008visualizing} visualization on EPIC.} (a) Clip-level features from \SC{I3D};  Node features for \SC{Ours} (b) without and (c) with GCN. Colors correspond to different kitchens. %
}
\label{fig:tsne}
\vspace{-4mm}
\end{figure}

%


Feature space visualizations show how clips for the same action (but different kitchens) cluster due to explicit label supervision (\reffig{fig:tsne}a), but kitchen-specific clusters arise naturally (\reffig{fig:tsne}c) in our method, encoding useful environment-aware information to improve performance.


\section{Conclusion}

We proposed a method to produce a topological affordance graph from egocentric video of human activity, highlighting commonly used zones that afford coherent actions across multiple kitchen environments.
Our experiments on scene affordance learning and long range anticipation demonstrate its viability as an enhanced representation of the environment gained from egocentric video.  Future work can leverage the environment affordances
to guide users in unfamiliar spaces with AR or allow robots to explore a new space through the lens of how it is likely used.

\vspace{0.1in}
\noindent\textbf{Acknowledgments}: Thanks to Jiaqi Guan for help with SLAM on EPIC, and Noureldien Hussein for help with the Timeception~\cite{hussein2019timeception} and Videograph~\cite{hussein2019videograph} models. UT Austin is supported in part by ONR PECASE and DARPA L2M.

{\small
\bibliographystyle{ieee}
\bibliography{egbib}
}

\newpage
\clearpage

\setcounter{section}{0}
\setcounter{figure}{0}
\setcounter{table}{0}
\renewcommand{\thesection}{S\arabic{section}}
\renewcommand{\thetable}{S\arabic{table}}
\renewcommand{\thefigure}{S\arabic{figure}}

\section*{Supplementary Material}

This section contains supplementary material to support the main paper text. The contents include:

\begin{itemize}[leftmargin=*]
\itemsep0em 
    \item (\S\ref{sec:supp_demo}) A video demonstrating our \SC{Ego-Topo} graph construction process following \refalg{alg:topomap}, and our scene affordance results from \refsec{sec:affordance_exps} in the main paper. 
    \item (\S\ref{sec:supp_annotation}) Setup and details for crowdsourced affordance annotation on EPIC and EGTEA+.
    \item (\S\ref{sec:supp_class_AP}) Class-level breakdown of affordance prediction results from \reftbl{tbl:affordance}.   
    \item (\S\ref{sec:supp_algo1}) Additional implementation details for the graph construction in \refsec{sec:Approach1}.
    \item (\S\ref{sec:supp_exps}) Implementation details for our models presented in \refsec{sec:Approach3} and \refsec{sec:Approach4}.
    \item (\S\ref{sec:supp_actionmaps}) Implementation details for \SC{ActionMaps} baseline from \refsec{sec:affordance_exps} (Baselines).
    \item (\S\ref{sec:supp_slam}) Implementation details for SLAM from \refsec{sec:affordance_exps} (Baselines).
    \item (\S\ref{sec:supp_qual}) Additional affordance prediction results to supplement \reffig{fig:afford_qual}.
\end{itemize}

\section{\SC{Ego-Topo} demonstration video} \label{sec:supp_demo}
We show examples of our graph construction process over time from egocentric videos following \refalg{alg:topomap} in the main paper. The end result is a topological map of the environment where nodes represent primary spatial zones of interaction, and edges represent commonly traversed paths between them. Further, the video demonstrates our affordance prediction results from \refsec{sec:affordance_exps} over the constructed topological graph. The video and interface to explore the topological graphs can be found on the \href{http://vision.cs.utexas.edu/projects/ego-topo/}{project page}.

\begin{figure*}[t]
\centering
\includegraphics[width=\textwidth]{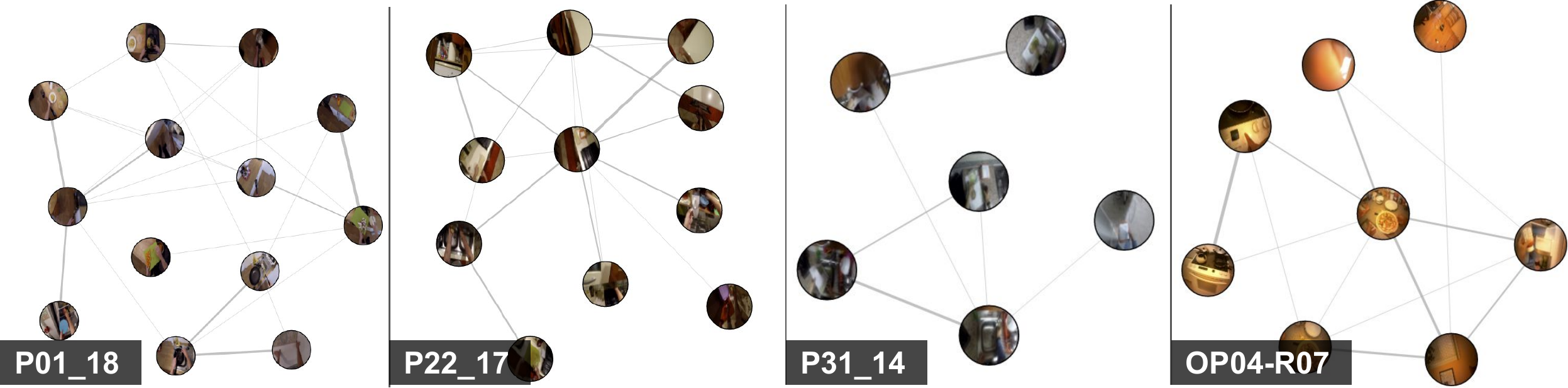}
\vspace*{-0.25in}
\caption{\textbf{\SC{Ego-Topo} graphs constructed directly from egocentric video.} Each panel shows the output of \refalg{alg:topomap} for videos in EPIC (panels 1-3) and EGTEA+ (panel 4). Connectivity represents frequent paths taken by humans while using the environment. Edge thickness represents how frequently they are traversed.
}
\vspace*{-0.10in}
\label{fig:supp_final_graphs}
\end{figure*}

\reffig{fig:supp_final_graphs} shows static examples of fully constructed topological maps from a single egocentric video from the test sets of EPIC and EGTEA+. %
Graphs built from long videos with repeated visits to nodes (P01\_18, P22\_07) result in a more complete picture of the environment. Short videos where only a few zones are visited (P31\_14) can be linked to other graphs of the same kitchen (\refsec{sec:Approach2}). The last panel shows a result on EGTEA+.

\section{%
Crowdsourced affordance annotations} \label{sec:supp_annotation}
As mentioned in \refsec{sec:affordance_exps}, we collect annotations for afforded interactions for EPIC and EGTEA+ video frames to evaluate our affordance learning methods. We present annotators with a single frame (center frame) from a video clip and ask them to select \emph{all likely} interactions that occur in the location presented in the clip. Note that these annotations are used exclusively for evaluating affordance models --- they are trained using single-clip interaction labels (See \refsec{sec:Approach3}).

\begin{figure*}[t]
\centering
\includegraphics[width=1.8\columnwidth]{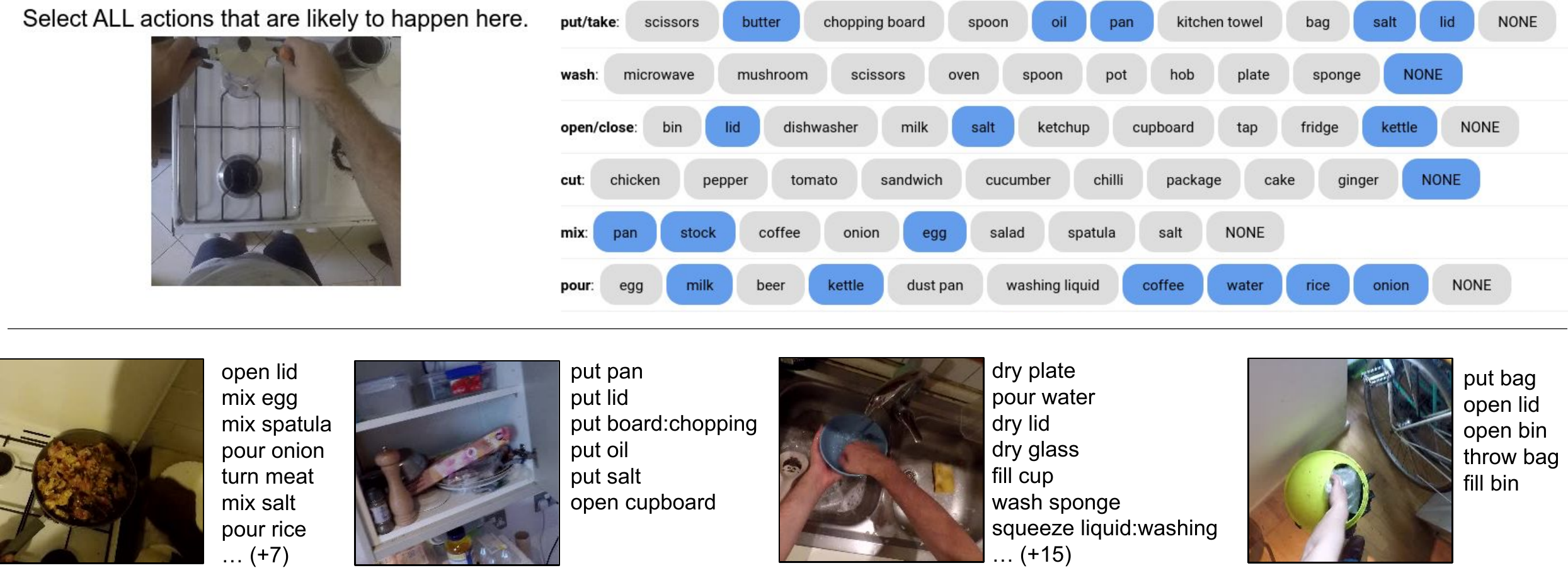}
\vspace*{-0.1in}
\caption{\textbf{Crowdsourcing affordance annotations.} \textbf{(Top panel)}  Affordance annotation interface. Users are asked to identify all \emph{likely} interactions at the given location. 6 out of 15 afforded actions are shown here. \textbf{(Bottom panel)} Example affordance annotations by Mechanical Turk annotators. Only annotations where 3+ workers agree are retained. }
\vspace*{-0.10in}
\label{fig:supp_interface}
\end{figure*}

On EPIC, we select 120 interactions (verb-noun pairs) over the 15 most frequent verbs %
and for common objects that afford multiple interactions. For EGTEA+, we select all 75 interactions provided by the dataset. A list of all these interactions is in \reftbl{tbl:supp_interaction_list}. 
Each image is labeled by 5 distinct annotators, and only labels that 3 or more annotators agree on are retained. This results in 1,020 images for EGTEA+ and 1,155 images for EPIC.
Our annotation interface is shown in  \reffig{fig:supp_interface} (top panel), and examples of resulting annotations are shown in \reffig{fig:supp_interface} (bottom panel).

\begin{table*}[t!]
\centering
\resizebox{\textwidth}{!}{
\begin{tabular}{|p{1.3cm}|p{30cm}|} \hline
\B{EPIC} &
put/take: pan, spoon, lid, board:chopping, bag, oil, salt, towel:kitchen, scissors, butter;
open/close: tap, cupboard, fridge, lid, bin, salt, kettle, milk, dishwasher, ketchup;
wash: plate, spoon, pot, sponge, hob, microwave, oven, scissors, mushroom;
cut: tomato, pepper, chicken, package, cucumber, chilli, ginger, sandwich, cake;
mix: pan, onion, spatula, salt, egg, salad, coffee, stock;
pour: pan:dust, onion, water, kettle, milk, rice, egg, coffee, liquid:washing, beer;
throw: onion, bag, bottle, tomato, box, coffee, towel:kitchen, paper, napkin;
dry: pan, plate, knife, lid, glass, fork, container, hob, maker:coffee;
turn-on/off: kettle, oven, machine:washing, light, maker:coffee, processor:food, switch, candle;
turn: pan, meat, kettle, hob, filter, sausage;
shake: pan, hand, pot, glass, bag, filter, jar, towel;
peel: lid, potato, carrot, peach, avocado, melon;
squeeze: sponge, tomato, liquid:washing, lemon, lime, cream;
press: bottle, garlic, dough, switch, button;
fill: pan, glass, cup, bin, bottle, kettle, squash
\\ \hline
\B{EGTEA+} & 
inspect/read: recipe; 
open: fridge, cabinet, condiment\_container, drawer, fridge\_drawer, bread\_container, dishwasher, cheese\_container, oil\_container; 
cut: tomato, cucumber, carrot, onion, bell\_pepper, lettuce, olive; 
turn-on: faucet; 
put: eating\_utensil, tomato, condiment\_container, cucumber, onion, plate, bowl, trash, bell\_pepper, cooking\_utensil, paper\_towel, bread, pan, lettuce, pot, seasoning\_container, cup, bread\_container, cutting\_board, sponge, cheese\_container, oil\_container, tomato\_container, cheese, pasta\_container, grocery\_bag, egg; 
operate: stove, microwave; 
move-around: eating\_utensil, bowl, bacon, pan, patty, pot; 
wash: eating\_utensil, bowl, pan, pot, hand, cutting\_board, strainer; 
spread: condiment; 
divide/pull-apart: onion, paper\_towel, lettuce; 
clean/wipe: counter; 
mix: mixture, pasta, egg; 
pour: condiment, oil, seasoning, water; 
compress: sandwich; 
crack: egg; 
squeeze: washing\_liquid
 \\ \hline
\end{tabular} 
}
\caption{List of afforded interactions annotated for EPIC and EGTEA+.}%
\label{tbl:supp_interaction_list}
\end{table*}

\section{Average precision per class for affordances} \label{sec:supp_class_AP}
\begin{figure*}[t]
\centering
\includegraphics[width=\textwidth]{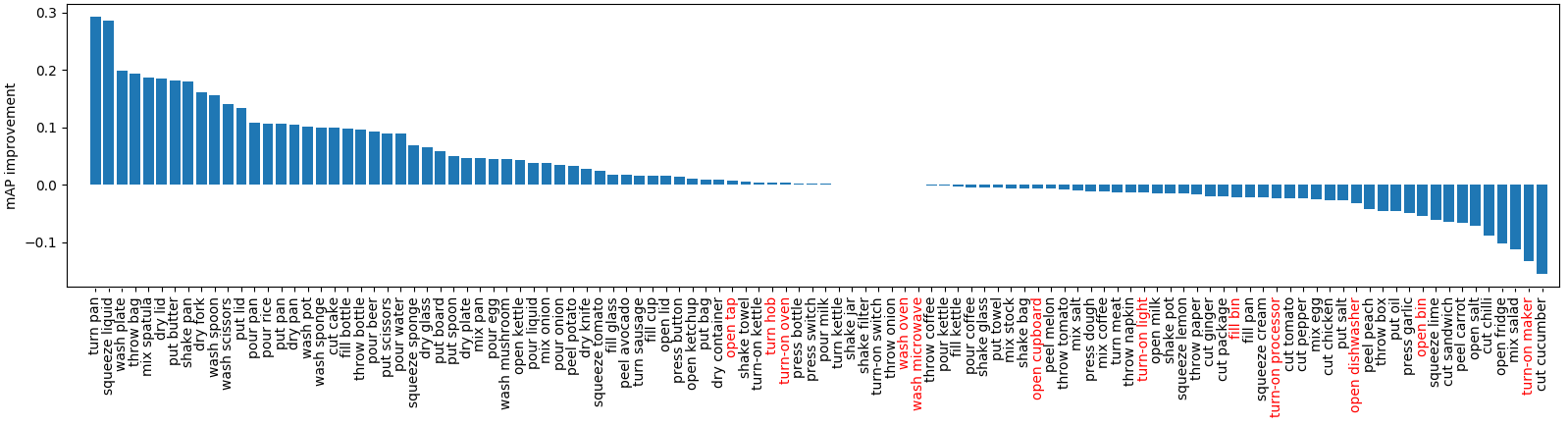}
\vspace*{-0.25in}
\caption{\textbf{Class-wise breakdown of average precision for affordance prediction on EPIC.} Our method outperforms the \SC{ClipAction} baseline on the majority of
classes. Single clip labels are sufficient for interactions that are strongly tied to a single physical location (red), whereas our method works particularly well for classes with objects that can be interacted with at multiple locations. 
}
\vspace*{-0.10in}
\label{fig:supp_class_dist}
\end{figure*}

As noted in our experiments in \refsec{sec:affordance_exps}, our method performs better on low-shot classes.  \reffig{fig:supp_class_dist} shows a class-wise breakdown of improvements achieved by our model over the \SC{ClipAction} model on the scene affordance task. Among the interactions, those involving objects that are typically tied to a single physical location, highlighted in red (\eg, fridges, stoves, taps etc.), are easy to predict, and do not improve much. Our method works especially well for interaction classes that occur in multiple locations (\eg, put/take spoons/butter, pour rice/egg etc.), which are linked in our topological graph.

\section{Additional implementation details for \SC{Ego-Topo} graph creation} \label{sec:supp_algo1}
We provide additional implementation details for our topological graph construction procedure from \refsec{sec:Approach1} and \refsec{sec:Approach2} in the main paper.

\noindent \textbf{Homography estimation details (\refsec{sec:Approach1}).} We generate SuperPoint keypoints~\cite{detone2018superpoint} using the pretrained model provided by the authors. For each pair of frames, we calculate the homography using 4 random points, and use RANSAC to maximize the number of inliers. We use inlier count as a measure of similarity.

\noindent \textbf{Similarity threshold and margin values in \refalg{alg:topomap} ($\sigma, m$).}  We fix our similarity threshold $\sigma = 0.7$ to ensure that only highly confident views are included in the graph. We select a large margin $m = 0.3$ to make sure that irrelevant views are readily ignored.  

\noindent \textbf{Node linking details (\refsec{sec:Approach2}).} We use hierarchical agglomerative clustering to link nodes across different environments based on functional similarity. We set the similarity threshold below which nodes will not be linked as 40\% of the average pairwise similarity between every node. We found that threshold values around this range (40-60\%) produced a similar number of clusters, while values beyond them resulted in too few nodes linked, or all nodes collapsing to a single node.

\noindent \textbf{Other details.} We subsample all videos to 6 fps. To calculate $s_f(f_t, n)$ in \refeqn{eq:sim}, %
we average scores for a window of 9 frames around the current frame, and we uniformly sample a set of 20 frames for each visit for robust score estimates. 

\section{Training details for affordance and long term anticipation experiments} \label{sec:supp_exps}
We next provide additional implementation and training details for our
experiments in \refsec{sec:exp} of the main paper.

\noindent \textbf{Affordance learning experiments in \refsec{sec:affordance_exps}.}
For all models, we use ImageNet pretrained ResNet-152 features for frame feature inputs. As mentioned in \refsec{sec:Approach3}, we use binary cross entropy (BCE) for our loss function. For original clips labeled with a single action label, we evaluate BCE for only the positive class, and mask out the loss contributions for all other classes. Adam with learning rate 1e-4,
weight decay 1e-6, and batch size 256 is used to optimize
the models parameters. All models are trained for 20 epochs, and learning rate is annealed once to 1e-5 after 15 epochs.

\noindent \textbf{Long term action anticipation experiments in \refsec{sec:anticipation_exps}.}
We pretrain an I3D model with ResNet-50 as the backbone on the original clip-level action recognition task for both EPIC-Kitchen and EGTEA+. Then, we extract the features from the pretrained I3D model for each set of 64 frames as the clip-level features. These features are used for all models in our long-term anticipation experiments.

Among the baselines, we implement \SC{TrainDist}, \SC{I3D}, \SC{RNN}, and \SC{ActionVlad}. For \SC{Timeception}, we import the authors' module\footnote{\url{https://github.com/noureldien/timeception}} and for \SC{Videograph}, we directly use the authors' implementation\footnote{\url{https://github.com/noureldien/videograph}} with our features as input.

For EPIC, all models are trained for 100 epochs with the learning rate starting from 1e-3 and decreased by a factor of 0.1 after 80 epochs. We use Adam as the optimization method with weight decay 1e-5 and batch size 256. For the smaller EGTEA+ dataset, we follow the same settings, except we train for 50 epochs.

\section{\SC{ActionMaps} implementation details} \label{sec:supp_actionmaps}
For the \SC{ActionMaps} method, we follow Rhinehart and Kitani~\cite{rhinehart2016learning} making a few necessary modifications for our setting. We use cosine similarity between pretrained ResNet-152 features to measure semantic similarity between locations as side information, instead of object and scene classifier scores, to be consistent with the other evaluated methods. We use the latent dimension 256 for the matrix factorization, and set $\lambda=\mu = 1e-3$ for the RWNMF optimization objective in~\cite{rhinehart2016learning}. We use location information in the similarity kernel only when it is available, falling back to just feature similarity when it is not (due to SLAM failures). We use this baseline in our experiments in \refsec{sec:affordance_exps}.

\section{SLAM implementation details} \label{sec:supp_slam}
We generate monocular SLAM trajectories for egocentric videos using the code and protocol from \cite{guan2019generative}. Specifically, we use ORB-SLAM2~\cite{mur2015orb} to extract trajectories for the full video, and drop timesteps where either tracking is unreliable or lost. We scale all trajectories by the maximum movement distance for each kitchen, so that (x, y) coordinates are bounded between [0, 1]. We create a uniform grid of squares, each with edge length 0.2 %
. We use this grid
to accumulate trajectories for the \SC{SLAM} baseline and to construct the \SC{ActionMaps} matrix in our experiments in \refsec{sec:affordance_exps}. We use the same process for EPIC and EGTEA+, with camera parameters from the dataset authors.

\begin{table}[]
\resizebox{\columnwidth}{!}{
\begin{tabular}{|l|l|l|}
\hline
                       &  EPIC (mAP) & EGTEA+ (mAP) \\ \hline
\SC{SLAM$_{5}$}            & 41.8        & 26.5         \\
\SC{SLAM$_{10}$}           & 41.3        & 26.5         \\
\SC{SLAM$_{20}$}           & 40.7        & 26.2         \\ \hline
\end{tabular}
}
\caption{\textbf{Affordance prediction results with varying grid sizes.} \SC{SLAM$_S$} refers to the \SC{SLAM} baseline from \refsec{sec:affordance_exps} with an $S \times S$ grid.}
\label{tbl:supp_grid_size}
\vspace{-0.1in}
\end{table}

We experimented with varying grid cell sizes (10x10, 20x20 grids), however, smaller grid cells resulted in very few trajectories registered to the same grid cell (\eg., for a 20x20 grid on EPIC, 61\% of cells register only a single trajectory) limiting the amount of labels that can be shared, and hence weakening the baseline. See \reftbl{tbl:supp_grid_size}.



\section{Additional node affordance results} \label{sec:supp_qual}
\begin{figure*}[t]
\centering
\includegraphics[width=\textwidth]{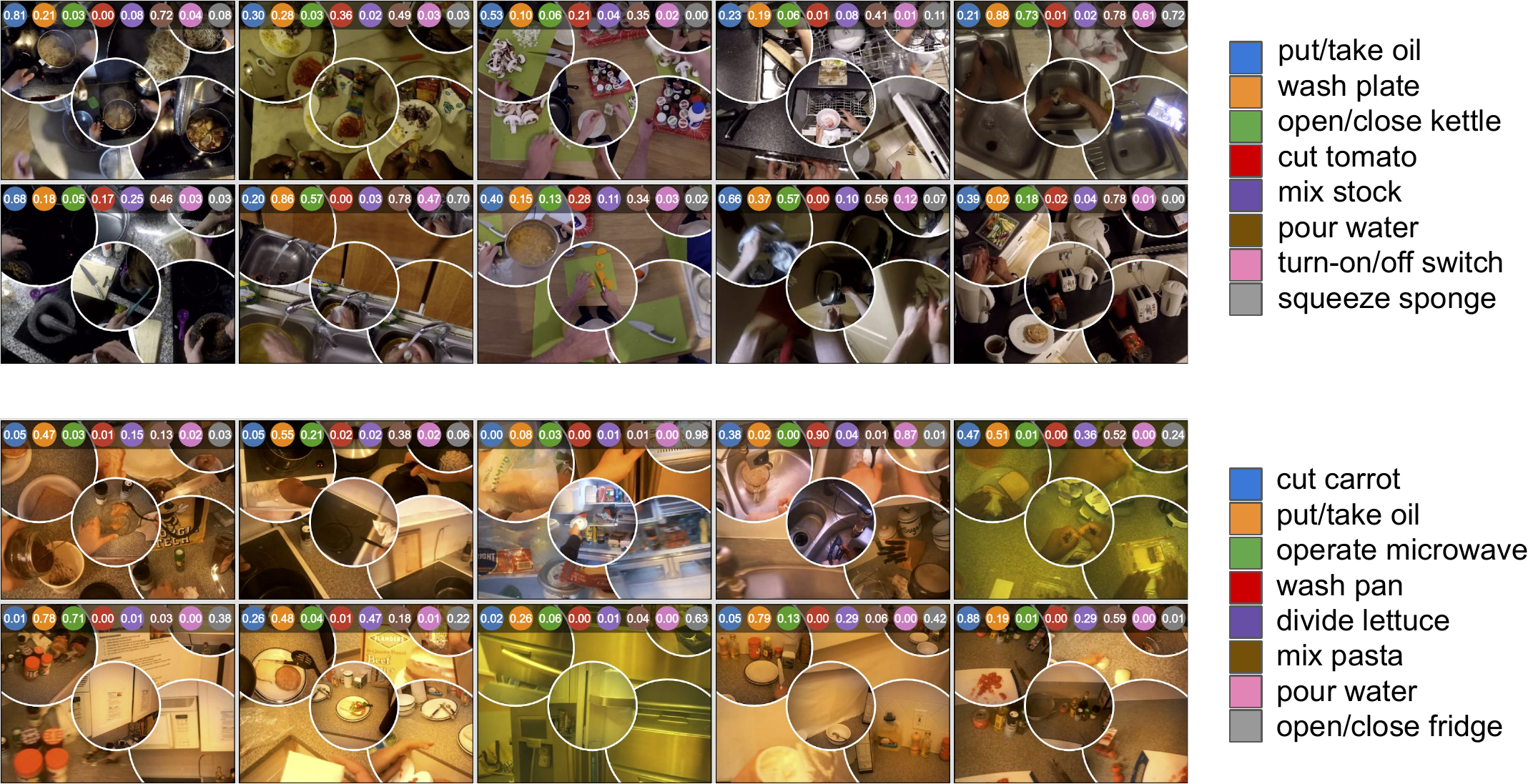}
\vspace*{-0.25in}
\caption{\textbf{Additional zone affordance prediction results.} Results on EPIC (top panel) and EGTEA+ (bottom panel).} %
\vspace*{-0.10in}
\label{fig:supp_afford_qual}
\end{figure*}

\reffig{fig:supp_afford_qual} provides more examples of affordance predictions by our model on zones (nodes) in our topological map, to supplement \reffig{fig:afford_qual} in the main paper. For clarity, we show 8 interactions on EPIC (top panel) and EGTEA+ (bottom panel), out of a total of 120 and 75 interactions respectively.   

\end{document}